\def\year{2016}\relax
\documentclass[letterpaper]{article}
\usepackage{aaai17}
\usepackage{url}
\usepackage{graphicx}
\usepackage{epstopdf}
\usepackage{amsfonts}
\usepackage{amssymb}
\usepackage{amsmath}
\usepackage{verbatim}
\usepackage{multirow}
\usepackage{subfigure}
\usepackage{array}
\usepackage{fp}
\usepackage{makecell}
\usepackage{flushend}
\usepackage{cases}
\usepackage{color}
\usepackage{times}
\usepackage{url}
\usepackage{latexsym}
\usepackage{graphicx}
\usepackage{epstopdf}
\usepackage{amsfonts}
\usepackage{amssymb}
\usepackage{amsmath}
\usepackage{verbatim}
\usepackage{multirow}
\usepackage{subfigure}
\usepackage{bbm}
\usepackage{times}
\usepackage{helvet}
\usepackage{courier}
\newcommand{\thickhline}{\noalign{\hrule height 1pt}}
\begin{document}
	%
	\title{Knowledge Enhanced Hybrid Neural Network for Text Matching}
		\author{
			Yu Wu$^\dag$\thanks{ The work was done when the first author was an intern in Microsoft Research Asia.}~~~~, Wei Wu$^\ddag$~~~~, Zhoujun Li$^\dag$~~~~, Ming Zhou$^\ddag$~~~~\\
			$^\dag$State Key Lab of Software Development Environment, Beihang University, Beijing, China\\
			$^\ddag$~~~~Microsoft Research, Beijing, China\\
			\{wuyu,lizj\}@buaa.edu.cn \{wuwei,mingzhou\}@microsoft.com 
		}
	\maketitle
	\begin{abstract}
		Long text brings a big challenge to semantic matching due to their complicated semantic and syntactic structures. To tackle the challenge, we consider using prior knowledge to help identify useful information and filter out noise to matching in long text. To this end, we propose a knowledge enhanced hybrid neural network (KEHNN). The model fuses prior knowledge into word representations by knowledge gates and establishes three matching channels with words, sequential structures of sentences given by Gated Recurrent Units (GRU), and knowledge enhanced representations. The three channels are processed by a convolutional neural network to generate high level features for matching, and the features are synthesized as a matching score by a multilayer perceptron. The model extends the existing methods by conducting matching on words, local structures of sentences, and global context of sentences. Evaluation results from extensive experiments on public data sets for question answering and conversation show that KEHNN can significantly outperform the-state-of-the-art matching models and particularly improve the performance on pairs with long text.
	\end{abstract}
	
	\section{Introduction}
	Semantic matching is a fundamental problem in many NLP tasks such as question answering (QA) \cite{voorhees1999trec}, conversation \cite{wang2013dataset}, and paraphrase identification \cite{dolan2004unsupervised}. Take question-answering as an example. Given a question and an answer passage, one can employ a matching function to measure their matching degree. The matching degree reflects how likely the passage can be used as an answer to the question.  
	
	The challenge of text matching lies in semantic gaps between natural language sentences. Existing work tackles the challenge by representing sentences or their semantic and syntactic relations from different levels of abstractions with neural networks \cite{hu2014convolutional,socher2011dynamic}.  These models only rely on the text within a pair to perform matching, whereas we find that sentences in a pair could have very complicated semantic and syntactic structures, and it is difficult for the-state-of-the-art neural models to extract useful features from such sentences to bridge the semantic gaps in the text pair.  Table \ref{example1} gives an example from community QA to illustrate the challenge. The answer is very long\footnote{The original answer has $149$ words.} and contains a lot of information that well compare the two schools but semantically far from the question (e.g., ``horse riding'' and ``lances swords'').  The information makes the answer a high quality one, but hinders the existing models from establishing the semantic relations between the question and the answer in matching. Similarly, when questions become long, matching also becomes difficult. In practice, such long text is not rare. For example, in a public QA data set, $54.8\%$ question answer pairs are longer than 60 words (question length plus answer length). More seriously, the-state-of-the-art model can only achieves  $74.2\%$ matching accuracy on pairs longer than 60 words compared to its performance $78.8\%$ on pairs shorter than 30 words. 
	These evidence indicates us that improving matching performance on pairs with long text is important but challenging, because the semantic gap is even bigger in such pairs.

	\begin{table}	
		\small
		\caption{A difficult example from QA \label{example1}}	
		\centering
		\begin{tabular}{m{8cm}}
			\hline
			\textbf{Question} : Which school is better Voltaire or Bonaparte? \\ \hline
			
			\textbf{Answer} : Both are good schools but
			Bonaparte will teach your kids to become a good leader but they concentrate mainly on outdoor physical activities, manoeuvers, strategies. Horse riding and lances swords are their speciality.... 			 		
			
			On the other hand Voltaire will make your child more of a philosopher! They encourage independent thinking...and mainly concentrates on indoor activities! They inculcate good moral values in the child and he will surely grow up to be a thinking person! \\ \hline
		\end{tabular}

	\end{table}	
	
	We study semantic matching in text pairs, and particularly, we aim to improve matching accuracy on long text. Our idea is that since it is difficult to establish the matching relations for pairs with long text only by themselves, we consider incorporating prior knowledge into the matching process. The prior knowledge could be topics, tags, and entities related to the text pair, and represents a kind of global context obtained elsewhere compared to local context such as phrases, syntactic elements obtained within the text in the pair. In matching, the global context can help filter out noise, and highlight parts that are important to matching. For instance, if we have a tag ``family'' indicating the category of the question in Table \ref{example1} in community QA,  we can use the tag to enhance the matching between the question and the answer. ``Family'' reflects the global semantics of the question. It strengthens the effect of its semantically similar words like ``kids'',``child'' and ``activity'' in QA matching, and at the same time reduce the influence of ``horse riding'' and ``lances swords'' to matching.  With the tag as a bridge, the semantic relation between the question and the answer can be identified, which is difficult to achieve only by themselves. 
	
	We propose a knowledge enhanced hybrid neural network (KEHNN) to leverage the prior knowledge in matching. 
	Given a text pair, KEHNN exploits a knowledge gate to fuse the semantic information carried by the prior knowledge into the representation of words and generates a knowledge enhanced representation for each word. The knowledge gate is a non-linear unit and controls how much information from the word is kept in the new representation and how much information from the prior knowledge flows to the representation. By this means, noise from the irrelevant words is filtered out, and useful information from the relevant words is strengthened. The model then forms three channels to perform matching from multiple perspectives. Each channel models the interaction of two pieces of text in a pair by a similarity matrix. The first channel matches text pairs on words. It calculates the similarity matrix by word embeddings. The second channel conducts matching on local structures of sentences. It captures sequential structures of sentences in the pair by a Bidirectional Recurrent Neural Network with Gated units (BiGRU) \cite{bahdanau2014neural}, and constructs the similarity matrix with the hidden vectors given by BiGRU. In the last channel, the knowledge enhanced representations, after processed by another BiGRU to further capture the sequential structures, are utilized to construct the similarity matrix. Since the prior knowledge represents global semantics of the text pair, the channel performs matching from a global context perspective. The three channels then exploit a convolutional neural networks (CNN) to extract compositional relations of the matching elements in the matrices as  high level features for matching. The features are finally synthesized as a matching score by a multilayer perceptron (MLP). The matching architecture lets two objects meet at the beginning, and measures their matching degree from multiple perspectives, thus the interaction of the two objects are sufficiently modeled.

	We conduct experiments on public data sets for QA and conversation. Evaluation results show that KEHNN can significantly outperform the-state-of-the-art matching methods, and particularly improve the matching accuracy on long text.

	Our contributions in this paper are three-folds: 1) proposal of leveraging prior knowledge to improve matching on long text; 2) proposal of a knowledge enhanced hybrid neural network which incorporates prior knowledge into matching in a general way and conducts matching on multiple levels; 3) empirical verification of the effectiveness of the proposed method on two public data sets.
	
	\section{Related Work}
	Early work on semantic matching is based on bag of words \cite{ramos2003using} and employs statistical techniques like LDA \cite{blei2003latent} and translation models \cite{koehn2003statistical} to overcome the semantic gaps. Recently,  neural networks have proven more effective on capturing semantics in text pairs. Existing methods can be categorized into two groups.  The first group follows a paradigm that matching is conducted by first representing sentences as vectors. Typical models in this group include DSSM \cite{huang2013learning}, NTN \cite{socher2013reasoning}, CDSSM \cite{shen2014latent}, Arc1 \cite{hu2014convolutional}, CNTN \cite{qiu2015convolutional}, and LSTMs \cite{tan2015lstm}. These methods, however, lose useful information in sentence representation, and leads to the emergence of methods in the second group. The second group matches text pairs by an interaction representation of sentences which allows them to meet at the first step. For example, MV-LSTM \cite{wan2015deep} generates the interaction representation by LSTMs and neural tensors, and then uses k-max pooling and a multi-layer perceptron to compute a matching score. MatchPymid \cite{pang2016text} employs CNN to extract features from a word similarity matrix. More effort along this line includes DeepMatch$_{topic}$ \cite{lu2013deep}, MultiGranCNN \cite{yin2015multigrancnn}, ABCNN \cite{yin2015abcnn}, Arc2 \cite{hu2014convolutional}, Match-SRNN \cite{wan2016match}, and Coupled-LSTM \cite{liu2016modelling}. Our method falls into the second group, and extends the existing methods by introducing prior knowledge into matching and conducting matching with multiple channels.

	\section{Approach}
	\subsection{Problem Formalization}
	Suppose that we have a data set $\mathcal{D} = \{(l_i,S_{x,i},S_{y,i})\}_{i=1}^N$, where $S_{x,i} = (w_0, \ldots, w_j, \ldots, w_I)$ and $S_{y,i} = (w_0^{'},\ldots, w_j^{'}, \ldots, w_J^{'})$ are two pieces of text, and $w_j$ and $w_j^{'}$ represent the $j$-th word of $S_{x,i}$ and $S_{y,i}$ respectively, and $N$ is the number of instances.  $l_i \in \{1,\ldots,C\}$ is a label indicating the matching degree between $S_{x,i}$ and $S_{y,i}$. In addition to $\mathcal{D}$, we have prior knowledge for $S_{x,i}$ and $S_{y,i}$ denoted as $\mathbf{k}_{x,i}$ and $\mathbf{k}_{y,i}$ respectively. Our goal is to learn a matching model $g(\cdot,\cdot)$  with $\mathcal{D}$ and $\left\lbrace \cup_{i=1}^N{ \mathbf{k}_{x,i}} , \cup_{i=1}^N{ \mathbf{k}_{y,i}} \right\rbrace $. Given a new pair $(S_{x},S_{y})$ with prior knowledge $(\mathbf{k}_{x},\mathbf{k}_{y})$, $g(S_{x},S_{y})$ predicts the matching degree between $S_{x}$ and $S_{y}$. 
	
	To learn $g(\cdot,\cdot)$, we need to answer two questions: 1) how to use prior knowledge in matching; 2) how to perform matching with both text pairs and prior knowledge.  In the following sections, we first describe our method on incorporating prior knowledge into matching, then we show details of our model. 
	
	\subsection{Knowledge Gate}
	%

	Inspired by the powerful gate mechanism \cite{hochreiter1997long,chung2014empirical} which controls information in and out when processing sequential data with recurrent neural networks (RNN), we propose using knowledge gates to incorporate prior knowledge into matching. The underlying motivation is that we want to use the prior knowledge to filter out noise and highlight the useful information to matching in a piece of text. Formally, let $e_w \in \mathbb{R}^d$ denote the embedding of a word $w$ in text $S_x$ and $\mathbf{k}_x \in \mathbb{R} ^ {n}$ denote the representation of the prior knowledge of $S_x$. Knowledge gate $k_w$ is defined as 
	\begin{equation}
	\label{knowledge_gate}
	k_w = \sigma(W_k e_w + U_k \mathbf{k}_x),
	\end{equation}
	where $\sigma$ is a sigmoid function, and $W_k \in \mathbb{R}^{d \times d}$, $U_k \in \mathbb{R}^{d \times n}$ are parameters.  
	With $k_w$, we define a knowledge enhanced representation for $w$ as
	\begin{equation}
	\widetilde{e}_w = k_w \odot e_w + (1-k_w) \odot \mathbf{k}_x, \label{eq:gate}
	\end{equation}
	where $\odot$ is an element-wise multiplication operation. Equation (\ref{eq:gate}) means that prior knowledge is fused into matching by a combination of the word representation and the knowledge representation. In the combination, the knowledge gate element-wisely controls how much information from word $w$ is preserved, and how much information from  prior knowledge $\mathbf{k}_x$ flows in. The advantage of the element-wise operation is that it offers a way to precisely control the contributions of prior knowledge and words in matching. Entries of $k_w$ lie in $[0,1]$. The larger an entry of $k_w$ is, the more information from the corresponding entry of $e_w$  will be kept in $\widetilde{e}_w$.  In contrast, the smaller an entry of $k_w$ is, the more information from the corresponding entry of $\mathbf{k}_x$ will flow into $\widetilde{e}_w$.  Since $k_w$ is determined by both $e_w$ and $\mathbf{k}_x$ and learned from training data, it will keep the useful parts in the representations of $w$ and the prior knowledge and at the same time filter out noise from them. 
	
	
	\subsection{Matching with Multiple Channels}
	\begin{figure*}[]		
		\begin{center}
			\includegraphics[width=16cm,height=5.5cm]{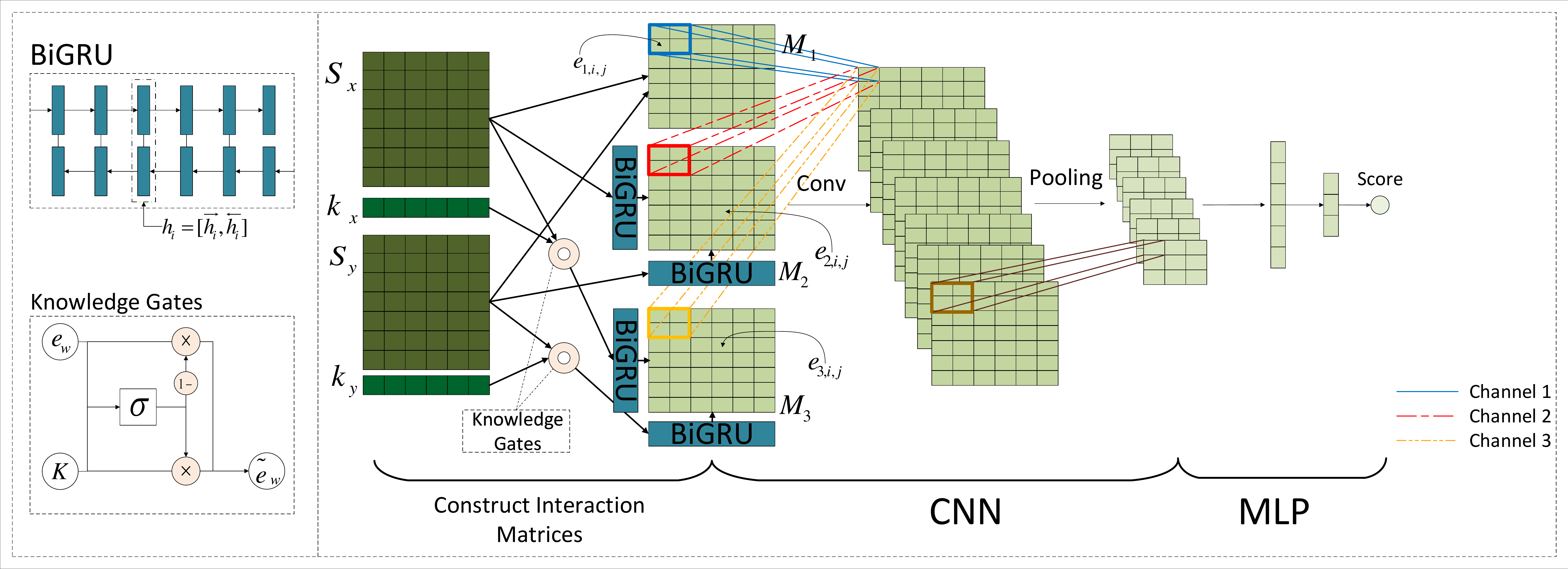}
		\end{center}
		
		\caption{Architecture of KEHNN}\label{fig:arch}
	\end{figure*}
	With the knowledge enhanced representations, we propose a knowledge enhanced hybrid neural network (KEHNN) which conducts matching with multiple channels. Figure \ref{fig:arch} gives the architecture of our model. Given a pair $(S_{x},S_{y})$, 
	the model looks up an embedding table and represents $S_{x}$ and $S_{y}$ as $\mathbf{S}_x = [e_{x,0}, \ldots, e_{x,i}, \ldots, e_{x,I}]$ and $\mathbf{S}_y = [e_{y,0}, \ldots, e_{y,i}, \ldots, e_{y,J}]$ respectively, where $e_{x,i}, e_{y,i} \in \mathbb{R} ^ d$ are the embeddings of the $i$-th word of $S_x$ and $S_y$ respectively. $\mathbf{S}_x$ and $\mathbf{S}_y$ are used to create three similarity matrices, each of which is regarded as an input channel of a convolutional neural network (CNN).  CNN extracts high level features from the similarity matrices. All features are finally concatenated and synthesized by a multilayer perceptron (MLP) to form a matching score.  
	
	Specifically, in channel one, $\forall i, j$, element $e_{1,i,j}$ in similarity matrix $\mathbf{M}_1$  is calculated by
	\begin{equation}\label{eq:f1}\small
	e_{1,i,j} = h(e_{x,i} ^{\intercal} \cdot e_{y,j}),
	\end{equation}
	where $h(\cdot)$ could be ReLU or tanh. $\mathbf{M}_1$ matches $S_{x}$ and $S_{y}$ on words.
	
	In channel two, we employ bidirectional gated recurrent units (BiGRU) \cite{chung2014empirical} to encode $S_x$ and $S_y$ into hidden vectors. A BiGRU consists of a forward RNN and a backward RNN. The forward RNN processes $S_x$ as it is ordered (i.e., from $e_{x,1}$ to $e_{x,I}$), and generates a sequence of hidden states $(\overrightarrow{h}_1, \ldots, \overrightarrow{h}_I)$. The backward RNN reads the sentence in its reverse order (i.e., from $e_{x,I}$ to $e_{x,1}$) and generates a sequence of backward hidden states $(\overleftarrow{h}_1, \ldots, \overleftarrow{h}_I)$. BiGRU then forms the hidden vectors of $S_x$ as $\{h_{x,i} = [\overrightarrow{h}_i ,\overleftarrow{h}_i]\}_{i=1}^I$ by concatenating the forward and the backward hidden states. More specifically, 
	$\forall i, \overrightarrow{h}_i \in \mathbb{R}^{m}$ is calculated by
	\begin{eqnarray}\small
	&& z_i = \sigma(W_z e_{x,i} + U_z  \overrightarrow{h}_{i-1}) \\
	&& r_i = \sigma(W_r  e_{x,i} + U_r  \overrightarrow{h}_{i-1}) \\
	&&　\widetilde{h}_i = tanh(W_h e_{x,i} + U_h (r_i \odot \overrightarrow{h}_{i-1}))　\\
	&&\overrightarrow{h}_i = z_i \odot \widetilde{h}_i + (1-z_i) \odot \overrightarrow{h}_{i-1},
	\end{eqnarray}
	where $z_i$ and $r_i$ are an update gate and a reset gate respectively, and $W_z$, $W_h$, $W_r$, $U_z$, $U_r$,$ U_h $ are parameters. The backward hidden state $\overleftarrow{h}_i \in \mathbb{R}^{m}$ is obtained in a similar way. Following the same procedure, we get $\{h_{y,i}\}_{i=1}^J$ as the hidden vectors of $S_y$ . With the hidden vectors, 
	$\forall i, j$, we calculate element $e_{2,i,j}$ in similarity matrix $\mathbf{M}_2$  by
	\begin{equation}\label{eq:f2}\small
	e_{2,i,j} = h(h_{x,i} ^{\intercal} W_2 h_{y,j} + b_2),
	\end{equation}
	where $W_2 \in \mathbb{R}^{2m \times 2m}$ and $b_2 \in \mathbb{R}$ are parameters. Since BiGRU encodes sequential information of sentences into hidden vectors, $\mathbf{M}_2$ matches $S_x$ and $S_y$ on local structures (i.e., sequential structures) of sentences.    
	
	
	In the last channel, we employ another BiGRU to process the sequences of $S_x$ and $S_y$ which consists of the knowledge enhanced representations in Equation (\ref{eq:gate}), and obtain the knowledge enhanced hidden states  $\mathbf{kh}_x = (kh_{x,1},\ldots, kh_{x,I})$ and $\mathbf{kh}_y = (kh_{y,1}, \ldots, kh_{y,J})$ for $S_x$ and $S_y$ respectively. Similar to channel two, $\forall i,j$, element $e_{3,i,j}$ in similarity matrix $\mathbf{M}_3$ is given by
	\begin{equation}\label{eq:f3}\small
	e_{3,i,j} = h(kh_{x,i} ^{\intercal} \cdot W_3 \cdot kh_{y,j} + b_3),
	\end{equation}
	where $W_3 \in \mathbb{R}^{2m \times 2m}$ and $b_3 \in \mathbb{R}$ are parameters. Prior knowledge represents a kind of global semantics of $S_x$ and $S_y$, and therefore $\mathbf{M_3}$ matches $S_x$ and $S_y$ on global context of sentences. 
	
	The similarity matrices are then processed by a CNN to abstract high level features. $\forall i=1,2,3$, CNN regards a similarity matrix as an input channel, and alternates convolution and max-pooling operations. Suppose that $z^{(l,f)} = \left[ z^{(l,f)}_{i,j}  \right]_{I^{(l,f)} \times J^{(l,f)}}$ denotes the output of  feature maps of type-$f$ on layer-$l$, where $z^{(0,f)}= \mathbf{M}_f$, $\forall f = 1,2,3$. On convolution layers (i.e. $\forall l = 1,3,5,\ldots,$), we employ a 2D convolution operation with a window size ${r_w^{(l,f)} \times r_h^{(l,f)}}$, and define $z_{i,j}^{(l,f)}$ as
	\begin{equation}\small
	z_{i,j}^{(l,f)} = \sigma (\sum_{f'=0}^{F_{l-1}} \sum_{s=0}^{r_w^{(l,f)}} \sum_{t=0}^{r_h^{(l,f)}} \mathbf{w}_{s,t} ^ {(l,f)} \cdot z_{i+s,j+t} ^{(l-1,f')} + b^{l,k} ), 
	\end{equation}
	where $\sigma(\cdot)$ is a ReLU, and  $\mathbf{w} ^ {(l,f)} \in \mathbb{R}^{r_w^{(l,f)} \times r_h^{(l,f)}} $ and $b^{l,k}$ are parameters of the $f$-th feature map on the $l$-th layer, and $F_{l-1}$ is the number of feature maps on the $(l-1)$-th layer. An max pooling operation follows a convolution operation and can be formulated as 
	\begin{equation}\small
	z_{i,j}^{(l,f)} = \max_{p_w^{(l,f)} >  s \geq 0} \max_{p_h^{(l,f)} > t \geq 0} z_{i+s,j+t} , \forall l = 2,4,6,\ldots,
	\end{equation}
	where $p_w^{(l,f)}$ and $p_h^{(l,f)}$ are the width and the height of the 2D pooling respectively. 
	
	The output of the final feature maps are concatenated as a vector $v$ and fed to a two-layer feed-forward neural network (i.e., MLP) to calculate a matching score $g(S_x, S_y)$:
	
	\begin{equation}\small
	g(S_{x},S_{y}) = \sigma_1\left(\mathbf{w}_2 ^{\intercal} \cdot\sigma_2\left({\mathbf{w}_1 ^{\intercal}  v +b_4}\right)+b_5\right),
	\end{equation}
	where $\mathbf{w}_1$, $\mathbf{w}_2$, $b_4$, and $b_5$ are parameters. $\sigma_1(\cdot)$ is softmax and $\sigma_2(\cdot)$ is tanh.
	
	KEHNN inherits the advantage of 2D CNN \cite{pang2016text,wan2015deep} that matching two objects by letting them meet at the beginning. Moreover, it constructs interaction matrices by considering multiple matching features. Therefore semantic relations between the two objects can be sufficiently modeled and leveraged in building the matching function. Our model extends the existing models \cite{hu2014convolutional} by fusing extra knowledge into matching and conducting matching with multiple channels.  
	
	We learn $g(\cdot, \cdot)$ by minimizing cross entropy \cite{levin1988accelerated} with $\mathcal{D}$ and $\left\lbrace \cup_{i=1}^N{ \mathbf{k}_{x,i}} , \cup_{i=1}^N{ \mathbf{k}_{y,i}} \right\rbrace$. Let $\Theta$ denote the parameters of our model. Then the objective function of learning can be formulated as  
	\begin{equation}\label{obj}\small
	\mathcal{L}(\mathcal{D};\Theta) = - \sum_{i=1}^{N}\sum_{c=1}^{C} P_c^g(l_i) \cdot log(P_c(g(S_{x,i},S_{y,i})),
	\end{equation}
	where $N$ in the number of instances in $\mathcal{D}$, and $C$ is the number of values of labels in $\mathcal{D}$. $P_c(g(S_{x,i},S_{y,i})$ returns the $c$-th element from the $C$-dimensional vector $g(S_{x,i},S_{y,i})$, and $P_c^g(l_i)$ is $1$ or $0$, indicating whether $l_i$ equals to $c$ or not. We optimize the objective function using back-propagation and the parameters are updated by stochastic gradient descent with Adam algorithm \cite{kingma2014adam}. As regularization, we employ early-stopping \cite{lawrence2000overfitting} and dropout \cite{srivastava2014dropout} with rate of 0.5. We set the initial learning rate and the batch size as $0.01$ and $50$ respectively.
	
	
	\subsection{Prior Knowledge Acquisition}
	Prior knowledge plays a key role to the success of our model. As described above, in learning, we expect prior knowledge to represent global context of input. In practice, we can use tags, keywords, topics, or entities that are related to the input as instantiation of the prior knowledge. Such prior knowledge could be obtained either from the metadata of the input, or from extra algorithms, and represent a summarization of the overall semantics of the input. Algorithms include tag recommendation \cite{wu2016improving}, keyword extraction \cite{wu2015mining}, topic modeling \cite{blei2003latent} and entity linking \cite{han2011collective} can be utilized to extract the prior knowledge from multiple resources like web documents, social media and knowledge base.    
	
	In our experiments, we use question categories as the prior knowledge in the QA task, because the categories assigned by the askers can reflect the question intention. For conversation task, we pre-trained a Twitter LDA model \cite{zhao2011comparing} on external large social media data, as the topics learning from social media could help us group text with similar meaning in a better way. Both the categories and the topics represent a high level abstraction from human or an automatic algorithm to the QA pairs or the message-response pairs, and therefore, they can reflect the global semantics of the input of the two tasks. As a consequence, our knowledge gate can learn a better representation for matching with the prior knowledge.
	
	\section{Experiments}
	We tested our model on two matching tasks: answer selection for question answering and response selection for conversation. 
	\subsection{Baseline}
	
	We considered the following models as baselines:
	
	%
	
	\textbf{Multi-layer perceptron (MLP)}: each sentence is represented as a vector by averaging its word vectors. The two vectors were fed to a two-layer feedforward neural network to calculate a matching score. MLP shared the embedding tables with our model.
	
	\textbf{DeepMatch$_{topic}$}: the matching model proposed in \cite{lu2013deep} which only used topic information to perform matching.
	
	\textbf{CNNs}: the Arc1 model and the Arc2 model proposed by Hu et al. \shortcite{hu2014convolutional}.  
	
	\textbf{CNTN}: the convolution neural tensor network \cite{qiu2015convolutional} proposed for community question answering. 
	
	\textbf{MatchPyramid}: the model proposed by Pang et al. \cite{pang2016text} who match two sentences using an approach of image recognition. The model is a special case of our model with only channel one.
	
	\textbf{LSTMs}: sentence vectors are generated by the last hidden state of LSTM \cite{lowe2015ubuntu}, or the attentive pooling result of all hidden states \cite{tan2015lstm}. We denote the two models as LSTM and LSTM$_a$.

	\textbf{MV-LSTM}:  the model \cite{wan2015deep} generates an interaction vector by combining hidden states of two sentences given by a shared BiLSTM. Then the interaction vector is fed to an MLP to compute the matching score.
	
	We implemented all baselines and KEHNN by an open-source deep learning framework Theano \cite{2016arXiv160502688short} . For all baselines and our model, we set the dimension of word embedding (i.e.,$d$) as $100$ and the maximum text length (i.e., $I$ and $J$) as $200$. In LSTMs, MV-LSTM, and BiGRU in our model, we set the dimension of hidden states as $100$ (i.e., $m$). We only used one convolution layer and one max-pooling layer in all CNN based models, because we found that the performance of the models did not get better with the number of layers increased. For Arc2, MatchPyramid, and KEHNN, we tuned the window size in convolution and pooling in $\{(2,2),(3,3)(4,4)\}$, and found that $(3,3)$ is the best choice. The number of feature maps is $8$.  For Arc1 and CNTN, we selected the window size from $\{2,3,4\}$ and set it as $3$. The number of feature maps is $200$. In MLP, we tuned the dimension of the hidden layer in $\{50,200,400,800\}$ and set it as $50$. $S_x$ and $S_y$ in KEHNN shared word embeddings, knowledge embeddings, parameters of BiGRUs, and parameters of the knowledge gates. All tuning was conducted on validation sets. The activation functions in baselines are the same as those in our model.
	\begin{table}[]
		\small
		\caption{Statistics of the answer selection data set \label{exp:qastat}}	
		\centering
		\begin{tabular}{c|c|c|c}
			\thickhline
			Data & $\#$question & $\#$answer & $\#$answers per question \\ \hline
			Training & 2600 & 16541 & 6.36\\
			Dev & 300 & 1645 & 5.48\\ 
			Test & 329 & 1976 & 6.00\\ 
			\thickhline
		\end{tabular}
	\end{table}
	
	\subsection{Answer Selection}
	The goal of answer selection is to recognize high quality answers in answer candidates of a question. We used a public data set of answer selection in SemEval 2015 \cite{alessandromoschitti2015semeval}, 
	which collects question-answer pairs from Qatar Living Forum\footnote{\url{http://www.qatarliving.com/forum}} and requires to classify the answers into $3$ categories (i.e. $C=3$ in our model) including good, potential and bad. The ratio of the three categories is $51 : 10: 39$. The statistics of the data set is summarized in Table \ref{exp:qastat}.
	We used classification accuracy as an evaluation metric.
	
	\subsubsection{Specific Setting}
	In this task, we regarded question categories tagged by askers as prior knowledge (both $\mathbf{k}_{x}$ and $\mathbf{k}_{y}$). There are $27$ categories in the Qatar Living data. Knowledge vector $\mathbf{k}$ was initialized by averaging the embeddings of words in the category. For all baselines and our model, the word embedding and the topic model (in DeepMatch$_{topic}$) were trained on a Qatar living raw text provided by SemEval-2015 \footnote{\url{http://alt.qcri.org/semeval2015/task3/index.php?id=data-and-tools}}. We fixed the word embedding during the training process, and set $h$ in Equation (\ref{eq:f1}), (\ref{eq:f2}), (\ref{eq:f3}) as ReLU.
	\begin{table}[]
		\small
		\caption{Evaluation results on answer selection \label{exp:answer}}	
		\centering
		\begin{tabular}{l|c}
			\thickhline
			& ACC  \\ \hline
			MLP  & 0.713\\ 
			DeepMatch$_{topic}$ & 0.682 \\
			Arc1  & 0.715\\ 
			Arc2  & 0.715\\ 
			CNTN  & 0.735\\ 
			MatchPyramid & 0.717\\
			LSTM  & 0.725\\ 
			LSTM$_a$ & 0.736 \\
			MV-LSTM  & 0.735\\
			KEHNN    & \textbf{0.748}\\
			JAIST  & 0.725\\  
			\thickhline
		\end{tabular}
	\end{table}
	\subsubsection{Results}
	JAIST, the champion of the task in SemEval15, used 12 features and an SVM classifier and achieved an accuracy of $0.725$. From Table \ref{exp:answer}, we can see that advanced neural networks, such as CNTN, MV-LSTM, LSTM$_a$ and KEHNN,  outperform JAIST's model, indicating that hand-crafted features are less powerful than deep learning methods.  Models that match text pairs by interaction representations like Arc2 and MatchPyramid are not better than models that perform matching with sentence embeddings like Arc1. This is because the training data is small and we fixed the word embedding in learning.   LSTM based models in general performs better than CNN based models, because they can capture sequential information in sentences. KEHNN leverages both the sequential information and the prior knowledge from categories in matching by a CNN with multiple channels. Therefore, it outperforms all other methods, and the improvement is statistically
	significant (t-test with p-value $\leq 0.05$). It is worthy to note that the gap between different methods is not big. This is because answers labeled as "potential" only cover $10\%$ of the data and are hard to predict.   
	
	\subsection{Response Selection}
	Response selection is important for building retrieval-based chatbots \cite{wang2013dataset}. The goal of the task is to select a proper response for a message from a candidate pool to realize human-machine conversation. We used a public English conversation data set, the Ubuntu Corpus \cite{lowe2015ubuntu}, to conduct the experiment. The corpus consists of a large number of human-human dialogue about Ubuntu technique. Each dialogue contains at least 3 turns, and we only kept the \textbf{last two utterances} as we study text pair matching and ignore context information. We used the data pre-processed by Xu et al. \cite{xu2016incorporating}\footnote{\url{https://www.dropbox.com/s/2fdn26rj6h9bpvl/ubuntu data.zip?dl=0}}, in which all urls and numbers were replaced by ``$\_url\_$'' and ``$\_number\_$'' respectively to alleviate the sparsity issue.  The training set contains $1$ million message-response pairs with a ratio $1 : 1$ between positive and negative responses, and both the validation set and the test set have $0.5$ million message-response pairs with a ratio
	$1 : 9$ between positive and negative responses. We followed Lowe et al. \cite{lowe2015ubuntu} and employed recall at position $k$ in $n$ candidates as evaluation metrics and denoted the metrics as $R_n @ k$.  $R_n @ k$ indicates if the correct response is in the top $k$ results from $n$ candidates.
	\subsubsection{Specific Setting}
	In this task, we trained a topic model to generate topics for both messages and responses as prior knowledge. We crawled 8 million questions (question and description) from the "Computers \& Internet" category in Yahoo! Answers, and utilized these data to train a Twitter LDA model \cite{zhao2011comparing} with $100$ topics. In order to construct $\mathbf{k}_{x,i}$ and $\mathbf{k}_{y,i}$, we separately assigned a topic to a message and a response by the inference algorithm of Twitter LDA. Then we transformed the topic to a vector by averaging the embeddings of top $20$ words under the topic. Word embedding tables were initialized using the public word vectors available at \url{http://nlp.stanford.edu/projects/glove} (trained on Twitter) and updated in learning. Tanh is used as $h$ in Equation (\ref{eq:f1}), (\ref{eq:f2}), (\ref{eq:f3}).

	\begin{table}[]
		\small
		\caption{Evaluation results on response selection \label{exp:response}}	
		\centering
		\begin{tabular}{l|c|c|c|c}
			\thickhline
			&  R$_2$@1      &  R$_{10}$@1 &  R$_{10}$@2&  R$_{10}$@5   \\ \hline
			MLP & 0.651 & 0.256 & 0.380 & 0.703 \\ 
			DeepMatch$_{topic}$ & 0.593 & 0.345 & 0.376 & 0.693\\ 
			Arc1 & 0.665 & 0.221 & 0.360 & 0.684\\ 
			Arc2 & 0.736 & 0.380 & 0.534 & 0.777\\ 
			CNTN & 0.743 & 0.349 & 0.512 & 0.797\\ 
			MatchPyramid & 0.743 & 0.420 & 0.554 & 0.786\\
			LSTM & 0.725 & 0.361 & 0.494 & 0.801\\ 
			LSTM$_a$ & 0.758 & 0.381 & 0.545 & 0.801\\ 
			MV-LSTM & 0.767 & 0.410& 0.565 &0.800\\ 
			KEHNN & \textbf{0.786} & \textbf{0.460} & \textbf{0.591} & \textbf{0.819}\\ 
			\thickhline
		\end{tabular}
	\end{table}
	\subsubsection{Results}
	Table \ref{exp:response} reports the evaluation results on response selection. Our method outperforms baseline models on all metrics, and the improvement is statistically significant (t-test with p-value $\leq 0.01$). In the data set, as the training data becomes large and we updated word embedding in learning, Arc2 and MatchPyraimd are much better than Arc1. LSTM based models perform better than CNN based models, which is consistent with the results in the QA task.  
	
	\subsection{Discussions}
	We first investigate the performance of KEHNN in terms of text length, as shown in Table \ref{exp:length}. We compared our model with 2 typical matching models: LSTM and MV-LSTM. We binned the text pairs into $4$ buckets, according to the length of the concatenation of the two pieces of text. $\#$Pair represents the number of pairs that fall into the bucket. From the results, we can see that on relatively short text (i.e., length in $[0,30)$ ), KEHNN performs comparably well with MV-LSTM, while on long text, KEHNN significantly improves the matching accuracy. The results verified our claim that matching with multiple channels and prior knowledge can enhance accuracy on long text. Note that on the Ubuntu data, all models perform worse on short text than them on long text. This is because we ignored context for short message-response pairs, while long pairs are usually independent with context and have complete semantics.  
	
	Furthermore, we also report the contributions of different channels of our model in Table \ref{exp:channel}. We can see that channel two is the most powerful one on the conversation data, while channel three is the best one on the QA data. This is because the prior knowledge in the conversation data is automatically generated  
	rather than obtained from meta-data like that in the QA data. The automatically generated prior knowledge contains noise which hurts the performance of channel three. The full model outperforms all single channels consistently, demonstrating that matching with multiple channels could leverage the three types of features and sufficiently model the semantic relations in text pairs. 
	\begin{table}[t]
		\small
		\caption{Accuracy on different length of text \label{exp:length}}	
		\centering
		\subtable[QA dataset]{
			\begin{tabular}{l|c|c|c|c}
				\thickhline
				Length &  $[0,30)$     &  $[30,60)$  & $[60,90)$ & $[90,\infty )$  \\ \hline
				$\#$Pair&   203  &  689 & 579 & 505  \\ 
				LSTM & 0.768 & 0.705 & 0.728 & 0.726  \\
				MV-LSTM & 0.788 & 0.708 & 0.746 & 0.739 \\
				KEHNN &0.792 & 0.721 & 0.765 & 0.746 \\ \hline
				\thickhline
			\end{tabular}
		}
		\subtable[Ubuntu dataset]{
			\begin{tabular}{l|c|c|c|c}
				\thickhline
				Length &  $[0,30)$     &  $[30,60)$  & $[60,90)$ & $[90,\infty )$  \\ \hline
				$\#$Pair&   253578   & 207772 & 33618  & 5032  \\ 
				LSTM & 0.707 & 0.748 & 0.732 & 0.718  \\
				MV-LSTM & 0.726 & 0.752 & 0.725 & 0.694 \\
				KEHNN & 0.724 & 0.774 & 0.785 & 0.791 \\ \hline
				\thickhline
			\end{tabular}
		}
		
	\end{table}
		
	\begin{table}[]	\vspace{-5mm}
		\small
		\caption{Comparison of different channels \label{exp:channel}}	
		\centering

		\begin{tabular}{l|c|c|c|c|c}
			\thickhline
			&   \multicolumn{4}{c|}{Conversation}    &        QA    \\ \hline
			&  R$_2$@1      &  R$_{10}$@1 &  R$_{10}$@2&  R$_{10}$@5 & ACC   \\ \hline
			only	$\mathbf{M}_1$ & 0.743 & 0.420 & 0.554 & 0.786 & 0.717\\
			only	$\mathbf{M}_2$ & 0.779 & 0.425 & 0.565 & 0.800 & 0.734\\
			only $\mathbf{M}_3$ & 0.750 & 0.360 & 0.531 & 0.791 & 0.738 \\ 
			KEHNN & 0.786 & 0.460 & 0.591 & 0.819 & 0.748\\
			\thickhline
			%
			
		\end{tabular}
		
	\end{table}
	\section{Conclusion}
	This paper proposed KEHNN that can leverages prior knowledge in semantic matching. Experimental results show that our model can significantly outperform state-of-the-art matching models on two matching tasks.
	\small
	\bibliographystyle {aaai17}
	\bibliography {aaai}
	
\end{document}